\documentclass[10pt,twocolumn,letterpaper]{article}

\usepackage{iccv}
\usepackage{times}
\usepackage{epsfig}
\usepackage{graphicx}
\usepackage{amsmath}
\usepackage{amssymb}
\usepackage{makecell,multirow,diagbox}

\usepackage[breaklinks=true,bookmarks=false]{hyperref}

\iccvfinalcopy 

\def\iccvPaperID{****} 

\begin{document}

\title{Utilizing Complex-valued Network for Learning to Compare Image Patches}

\author{Siwen Jiang,  Wenxuan Wei,  Shihao Guo, Hongguang Fu, Lei Huang\\
	School of Computer Science \& Engineering, UESTC, China\\
	{\tt\small jsw950515@gmail.com    \{weiwenxuan2711, jxjagsh, f\_hongguang\}@163.com mathhl@sina.com}
}

\maketitle

\begin{abstract}
	At present, the great achievements of convolutional neural network(CNN) in feature and metric learning have attracted many researchers. However, the vast majority of deep network architectures have been used to represent based on real values. The research of complex-valued networks is seldom concerned due to the absence of effective models and suitable distance of complex-valued vector.
	
	Motived by recent works, complex vectors have been shown to have a richer representational capacity and efficient complex blocks have been reported, we propose a new approach for learning image descriptors with complex numbers to compare image patches. We also propose a new architecture to learn image similarity function directly based on complex-valued network. We show that our models can perform competitive results on benchmark datasets. We make the source code of our models publicly available.
	
\end{abstract}

\section{Introduction}

One of the fundamental tasks in the field of computer vision is image matching, which plays an important role in some large tasks, such as image retrieval \cite{27}, robot navigation \cite{111} and texture classification \cite{29}. The general idea explored here is to learn a embedding vector for each image by using deep neural network, especially, the deep convolutional neural network \cite{16}. Traditionally, siamese architecture are widely adopted \cite{22,15,16,18,19}. The principle of this method is shown in Figure 1, the positive distance (D1) between the descriptor of similar images is reduced, while the negative distances (D2, D3) between descriptor of different images are enlarged, by optimizing the weights of the siamese network. In this way, different images can be dispersed to the greatest extent in the embedding space. And some other tasks can be also accomplished by the descriptor, including image clustering, face verification and semantic similarity.

Although deep learning subverts traditional methods in the tasks of computer vision such as target detection \cite{22}, image classification \cite{21} and face recognition \cite{11},  the 
\begin{figure}[t]
	\begin{center}
		\includegraphics[width=0.8\linewidth]{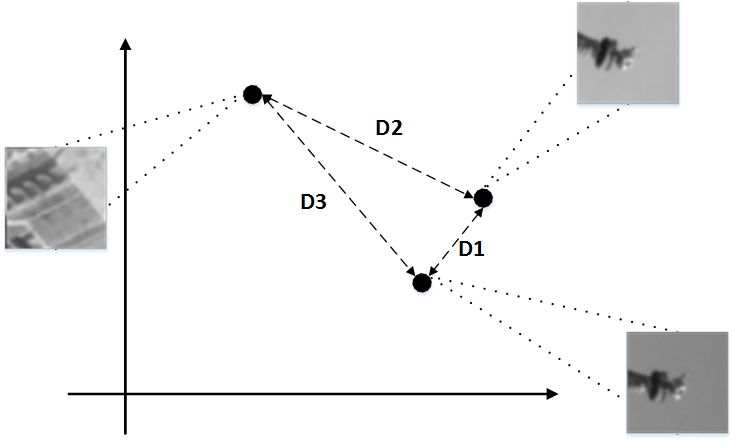}
	\end{center}
	\caption{Three different images are encoded and projected into the embedding space by triple net, two of which are similar. D1 denotes the embedded distance between them. D2 and D3 represent the embedded distance between another image and them, respectively}
	\label{fig:long}
	\label{fig:onecol}
\end{figure}
majority successful approaches focus on the exploration of the real number field, including compressing models \cite{23}, improving architecture and perfecting loss \cite{24}. 

With the reported of complex blocks \cite{10}, the new solutions for the problem of the initialization, the batch normalization \cite{25} and the activation function of complex network are available. Deep learning based on complex-valued are increasingly focused by researchers.

According to \cite{1,222,3,4,5}, there is a potential in complex network to enable easier learning, better generalization characteristics and to allow for preservation of detached detail. In the encoding network based on real-valued, details of the image may be lost due to improper length of descriptor. But in the complex-valued network, the phase value of descriptor vector can help the decoder recover it. Recent works have shown that, phase value provides a detailed description of objects because it contains the information of shapes, edges,and orientations of images \cite{10}.

In order to make full use of the advantage of complex-valued feature representation in the complex field, we propose the distance metric of the complex vector, improve the triple loss for the complex numbers, and apply it to the learning of image matching and descriptor. Our main contributions in this paper are as follows: 

1.We propose the complex triple network, which is used to extract complex descriptors.

2.We utilize 2-ch net\cite{2} to process the real part and the imaginary part of complex feature.

3.We present a new metric formula of the complex-valued embedding distance, which is referenced to the 	absolute distance.

4.We achieve a competitive performance result in the task of image patches comparing on the Photo-Tour and HPatches dataset.

\section{Related Work}

In the past, the primary approach to patches matching was using hand-crafted descriptors and comparing the squared euclidean distance, such as SIFT \cite{26}. The original method of using CNN to extract features had many problems. But with the development of neural network in recent years, new architectures have been emerging, such as AlexNet \cite{30}, VggNet \cite{31} and ResNet \cite{21}. At the same time, the method of data processing in deep learning are gradually improved, such as dropout \cite{13} and batch normalization \cite{25}. Utilizing CNN for feature extraction has become the major approach of image patches matching.

As a branch of the development of convolutional network, siamese network is a common method for descriptor learning and patches comparison. In 1993, LeCun and Y et al. utilized siamese networks for signature verification on American checks, that is, to verify whether the signature on checks is consistent with the bank's reserved signature \cite{14}. This is believed to be the first appearance of siamese networks. Similarly, siamese networks were utilized for face recognition \cite{16}. The descriptors of face images can be used to match new samples of previously unknown categories, by comparing the euclidean distances between image descriptors. However, because of the difficulty of descriptor learning in training, in the paper \cite{2}, the author proposed 2-ch net. Instead of comparing the distance by descriptors, it learns the image similarity function to match image patches through the neural network 

In the paper \cite{15,17}, the triplet network is proposed to change the twice branches structure into three branches. Its advantage lies in the distinction of details, that is, if input  images are similar, triple net can be used to describe the details better. The idea of triple net equivalent to adding two measures of input differences and learning a better representation of the inputs. Triplet net is also utilized for face recognition due to its advantage in descriptor learning \cite{11}. In further work, the author of \cite{19} improved the loss function of triple net, named PNSoft Loss.

Similar to siamese networks, complex networks also started in the 1990s. The theory and error propagation mechanism of complex networks has been explored by researchers \cite{6,7,8,9}. In the complex number field, the characteristics of data representation are more easily to abstracted. In some tasks, the complex-valued network has achieved significant advantages \cite{4,5,10}. Although the research has shown that the development potential and expandable space of complex networks, the development of complex networks has been marginalized because the training tricks of complex-valued networks need to be improved. In addition, complex networks require longer training time than the network based on real-value numbers. Recently, c. Trabelsi et al. proposed a solution for initialization, activation, and batch normalization in complex networks \cite{10}, which greatly simplified the training difficulty of complex-valued network.

Based on the above works, we propose a new approach to utilize complex network for learning to compare image patches. However, previous complex-valued networks cannot be used for descriptor learning because there is no suitable measurement method to describe the distance between complex vectors. Therefore, in this paper, we propose the measurement formula of complex vector. We present two architectures for matching image patches:

1.Complex Channel Net: learning to compare image patches directly

2.Complex Triple Net: learning to compare image patches via descriptors

\section{Architecture}
For the representation problem of complex-valued vector, we propose two solutions. The first is to output the degree of image similarity via learning a similarity function, called complex channel net. The Second is to learn the complex-valued descriptor of images, called complex triple net. The training and test approaches are different in these two ways. 

In training step, the input of complex channel net is a pair of image patches each time, and complex triple net is a triple of image patches. 

In the test step, complex channel net outputs the index of similarity via a pair of images directly. The complex triple net gets the descriptor for each image, and then compares the distance between them to get the result. We refer to the absolute value distance of the real number field and propose the distance of complex-valued vector. We introduced these in section 3.4 and 3.5 respectively.

In addition, the three modules that constitute the two structures names complex feature module, complex decision module and complex metric module, respectively. In section 3.1, we introduce complex feature module, which consist of three complex-valued blocks\cite{10}. The two networks we proposed all utilized it as the underlying block. In section 3.2, we propose the decision module. We utilize it to combine real and imaginary parts of complex-valued vector. In section 3.3, we introduce the metric module and the formula for measuring the distances of different complex-valued vectors in detail.

\begin{figure}[t]
	\begin{center}
		\includegraphics[width=0.8\linewidth]{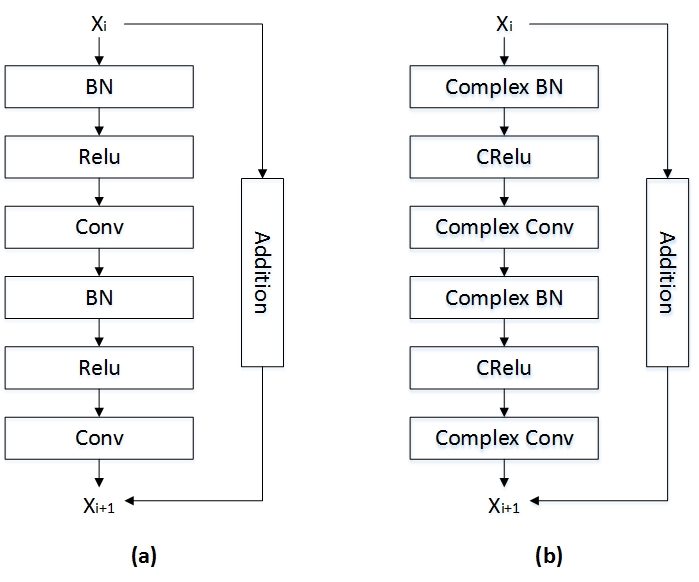}
	\end{center}
	\caption{Two basic blocks: (a) shows the real-valued residual blocks and (b) shows the complex-valued blocks}
	\label{fig:long}
	\label{fig:onecol}
\end{figure}
\subsection{Complex Feature Module}
In the paper \cite{21}, He et al. proposed a deep residual network, which allows the network to increase the number of layers and avoid degradation by means of identity mappings. It has been widely used in various network frameworks. The traditional residual block is shown on the Figure 2(a). The output of this block can be expressed as:
\begin{equation}X_{i+1}=\mathit{F}(X_{i},W_{i})+X_{i}\end{equation}
Where \(\mathit{F}(X_{i},W_{i})\) represent the operations present within a single real-valued residual block

Refer to the residual blocks of the real number to provide ideas for building a complex-valued residual blocks. We utilized the blocks proposed in \cite{10} and made some improvements. As shown in Figure 2(b), the structure of complex-valued blocks are the same as that of real numbers. The difference is that the operations of BN, Relu and Conv in them are based on complex number. The specific approaches is as follows:
\[BN\rightarrow Relu\rightarrow Conv\rightarrow BN\rightarrow Relu\rightarrow Conv\]
and \(\mathit{W_{i}}\)represents the parameters of the network to be learned.

\subsubsection*{a. Complex BN Layer}

In the real field, the formula for the BN algorithm is:
\begin{equation}
BN\left ( \tilde{x} \right )=\gamma \cdot x+\beta 
\end{equation}
Where \(\gamma=\sqrt{Var\left ( x \right )}\) and  \(\beta=E\left [ x \right ]\). \(\tilde{x}\) is the result of the normalization of the current sample \(\mathit{x}\) in the entire batch \(\mathit{X}\), represented as  \(\tilde{x}=Norm(\mathit{x}, \mathit{X})\). In \cite{10}, the author used the covariance matrix V to calculate it, and the formula is as follows:

\begin{equation}
V=\begin{pmatrix}
Cov(\Re (x),\Re (x)) & Cov(\Re (x),\Im (x)) & \\ 
Cov(\Re (x),\Im (x)) & Cov(\Im (x),\Im (x)) & 
\end{pmatrix}
\end{equation}

\begin{equation}
\tilde x = V^{-\frac{1}{2}}\left ( x-\mathit E\left [ x \right ] \right )
\end{equation}

Where \(\Re (x)\) and \(\Im (x)\) represent the real and imaginary parts of the eigenvector \(\mathit x\),  respectively. Different from \cite{10}, in order to simplify the calculation procedure, we use the BN algorithm for the real part and the imaginary part respectively. The expression is:
\begin{equation}
Complex BN(\tilde x)=BN(\Re (x))+ iBN(\Im (x))
\end{equation}

The improved complex BN algorithm also achieved competitive performance.

\subsubsection*{b. CRelu Layer}
The comparison of different complex activation functions has been given in \cite{10}. The complex activation method we give in the CRelu layer is the optimal way in that paper. The output of complex-BN layer is separated into the real part and imaginary part. They are activated  respectively by the Relu function, which can be expressed as:

\begin{equation}
CRelu(x)=Relu(\Re (x))+iRelu(\Im (x))
\end{equation}

\subsubsection*{c. Complex Conv Layer}
In the complex conv layer, we give a complex convolution kernel \(W_{c}=A+iB\), where \(\mathit A\) and \(\mathit B\) are both complex-valued matrices. The complex-valued vector  \(h=x+iy\) where \(\mathit x\) and \(\mathit y\) are both real vector. We can give the complex convolution operation as:

\begin{equation}W_{c} \cdot h=(Ax-By)+i(Bx+Ay)	\end{equation}

We construct complex feature module by stacking complex-block. The input and output of this module are complex-valued. We construct complex feature module with three complex-blocks, which is the basis of complex channel net and complex triple net.
\begin{figure}[t]
	\begin{center}
		\includegraphics[width=0.8\linewidth]{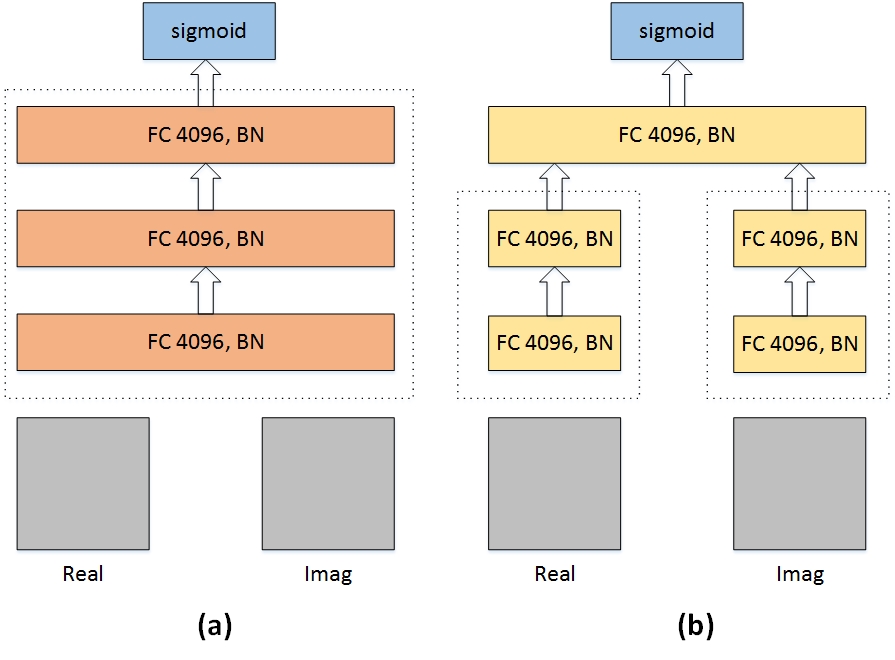}
	\end{center}
	\caption{Siamese net(a) and pesudo-siamese net(b). The difference between them is whether sharing the weights of network }
	\label{fig:long}
	\label{fig:onecol}
\end{figure}

\subsection{Decision Module}

Decision module at the top of complex channel net. In complex-valued feature module, each set of images will get the corresponding complex feature vectors.  We believe that these complex-valued vectors can express the similarity features of the two images. The decision module relies on this complex vectors to determine whether the input images are similar.

As shown in Figure 3, decision module separated the complex vectors into real and imaginary parts as the input of siamese network. According to the two forms of siamese network, we present the following two schemes:

\textbf{Siamese net: }As shown in Figure 3(a), the part of the dashed box is a siamese network. Real and Imag means the real part and the imaginary part of the feature vector. They are sent to the network respectively. After passing through the three fully connected layers where the weight is shared, the outputs of top layer are connected and sent to a single output fully connected layer with sigmoid activation function.

\textbf{Pesudo-siamese net:} Create two networks that do not share variables, as shown in Figure 3(b). Real and Imag are sent into the two networks in the dotted box respectively. Similar to the structure of siamese net, except the weights are not shared. Their outputs are connected and fed into the top level single output fully connected layer with sigmoid activation function.

In this paper, we use the siamese net. Unlike metric learning, decision module can learn a similarity function to output the index of similarity end to end. The advantage of this method is easier to train. But when classifying unknown samples, we need to compare all known samples by exhaustive method, which is very expensive.

\begin{figure}[t]
	\begin{center}
		\includegraphics[width=0.8\linewidth]{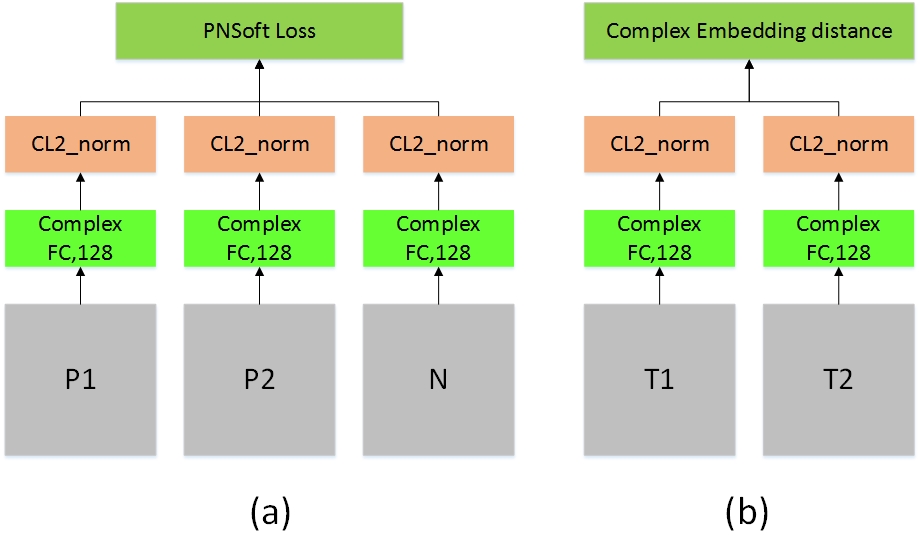}
	\end{center}
	\caption{The architecture of metric module. (a) represents the structure of the network in the training step and (b) represents the structure of the network in the test step.}
	\label{fig:long}
	\label{fig:onecol}
\end{figure}
\subsection{Metric Module}
Metric module at the top of complex triple net. Its main role is to learn descriptors from complex-valued eigenvectors, extracted by complex feature module.

In the training step, the input of complex triple net is a triple of images. Complex feature module is used to extract the complex-valued vector. The functions of metric module include dimensionality reduction, l2-normalization and loss optimization.

In the test step, metric module converts the complex feature vector extracted by complex feature module into complex descriptor. The image similarity can be determined, by comparing the distance between descriptors.

As shown in Fig 3(a), during the training step, three images are entered into the complex feature module, two of which are similar and the other is different. After the complex feature module, three complex vectors will be obtained, which are respectively P1, P2 and N. Where P1 and P2 are the complex-valued eigenvectors of two similar images, and N is the complex-valued eigenvector of the third image. 

There are two layers in metric module, including a complex fully connected layer and a complex l2-normalization layer. PNSoft loss has been used for descriptor learning\cite{19}. We use this loss function and change the euclidean distance in it to the distance of the complex-valued vector. Finally, image descriptors are learned by minimizing the proposed loss.

The author gives the following formula for the calculation of PNSoft Loss in \cite{19}:
\begin{equation}
L=[(\frac{e^{D(f_{p1},f_{p2})}}{e^{D(f_{p1},f_{p2})}+e^{D(f_{*})}})^{2}+(\frac{e^{D(f_{*})}}{e^{D(f_{p1},f_{p2})}+e^{D(f_{*})}})^{2}]
\end{equation}
Where \(f_{p1}\) and \(f_{p2}\) represents the output of \(p1\) and \(p2\) in metric module, namely the descriptor we learned. \(D(f_{p1},f_{p2})\) represents the euclidean distance between the descriptor \(f_{p1}\) and \(f_{p2}\). We will get three different distances, including a pairs of negative \(D(f_{p1},f_{pn})\), \(D(f_{p2},f_{pn})\) and a positive \(D(f_{p1},f_{p2})\). \(D(f_{*})\) is the minimum of \(D(f_{p1},f_{pn})\) and \(D(f_{p2},f_{pn})\).
\begin{figure*}
	\begin{center}
		\includegraphics[width=0.8\linewidth]{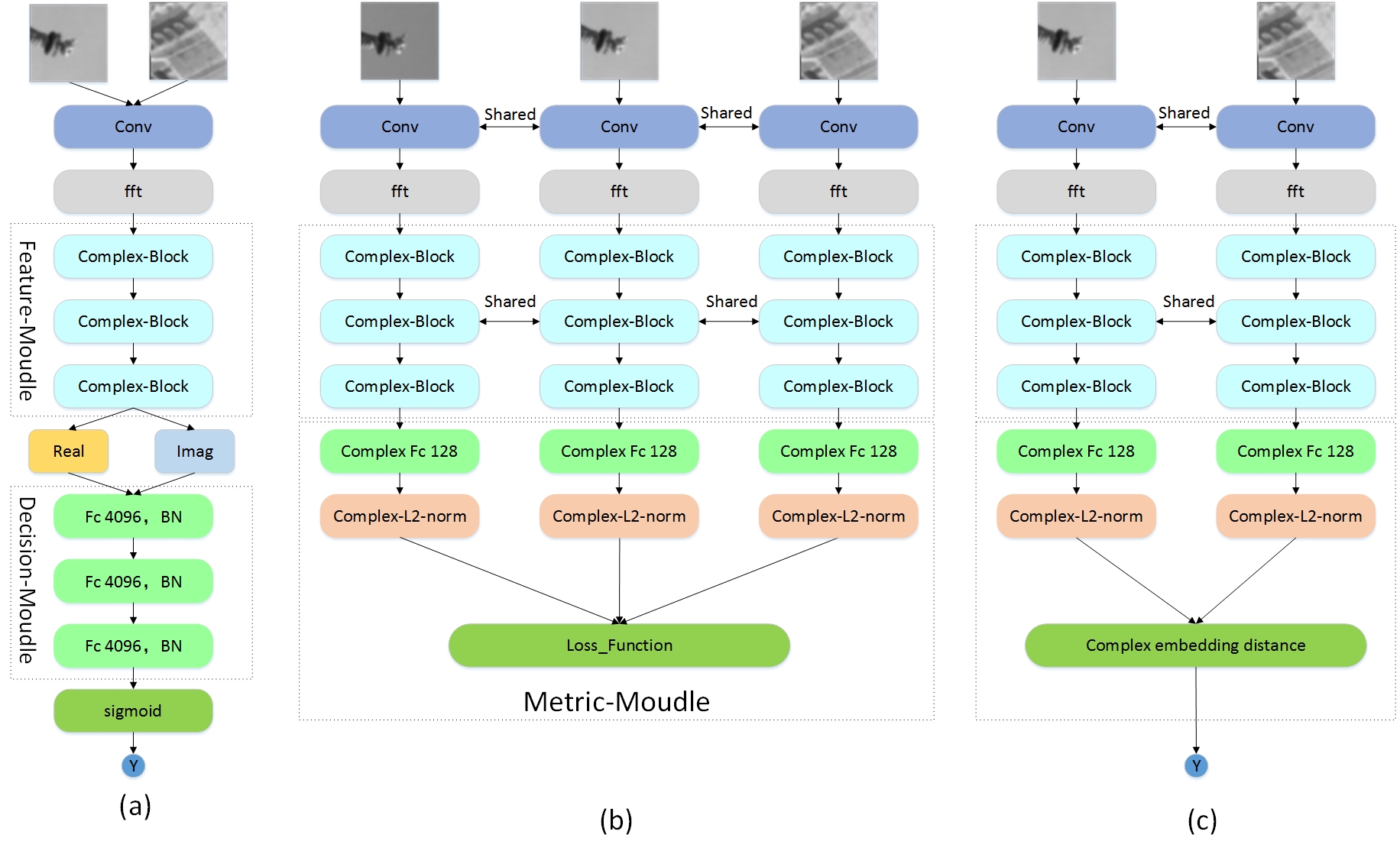}
	\end{center}
	\caption{Two proposed network architecture.(a) shows the structure of complex channel network. (b) shows the structure of complex triple net in the training step, and (c) shows the structure of complex triple net in the test step.}
	\label{fig:short}
\end{figure*}
This formula means that, among these three distances, the smallest negative distance should be greater than the positive distance. When the positive distance is 0, the first term of the formula is 0, and when the negative distance is infinity, the second term of the formula is 0. Through this formula, the positive example distance can be reduced and the negative example distance can be increased.

However, euclidean distance cannot describe the distance between complex vectors. Therefore, we modify the distance function D in PNSoft Loss, and propose the following formula to measure the distance between vectors in complex fields.
\begin{small}
	\begin{equation}
	D(f_{p1},f_{p2})=\sum_{i}^{n}\sqrt{[\Re (f_{p1}) - \Re(f_{p2})]^{2}-[\Im (f_{p1}) - \Im (f_{p2})]^{2}}
	\end{equation}
\end{small}
We also modify l2-normalization layer. When dealing with complex-valued features, the complex vector is separated into the real part and the imaginary part and treated separately. Our complex l2-normalized function is as follows:
\begin{equation}
Cl2\_norm(z)=L2\_norm(\Re (z)+iL2\_norm(\Im (z))
\end{equation}
where \(z \in C\).

In the test step, as shown in Figure 4(b), the images to be detected (T1, T2) are input into the bottom complex feature module. The outputs of metric module are the image descriptors based on the complex eigenvectors.

\subsection{Complex Channel Net}

The complex channel net consists of complex feature module and decision module. The main function of it is to learn a similarity function and directly output whether the input images are similar. 

As shown in Figure 5(a), its input is a pair of images connected together, also known as 2-channel image \cite{2}. After the first convolutional layer, the 2-channel image is converted into multiple channels feature maps. We convert it to a complex-valued form by the Fast Fourier Transform (FFT) and send it to the complex feature module.

The complex feature module in complex channel net consists of three complex blocks, consistent with the previous description, the output of it is divided into real parts and imaginary parts as input to the decision module. We use siamese network in decision module to process the real parts and imaginary parts, respectively, and connect their output as input to the full-connection layer with single output at the top of the network, activated with the sigmoid function.

In complex feature module, we add the max-pooling layer after each complex-block. Instead of pooling according to modulus length and amplitude, we separate the real part from the imaginary part. It can express as:
\begin{equation}
complex\_pool(z)=pool(\Re z)+ipool(\Im z)
\end{equation}

\begin{table*}
	\centering
	
	\begin{tabular}{c|c|c|c|c|c|c|c}
		\hline
		
		Training & Notredame & Liberty & Notredame&Yosemite& Yosemite&Liberty& \\
		\hline
		Test & \multicolumn{2}{c|}{Yosemite} &\multicolumn{2}{c|}{Liberty} &\multicolumn{2}{c|}{Notredame}&mean \\
		\hline
		SIFT\cite{26} & \multicolumn{2}{|c|}{27.29}& \multicolumn{2}{|c|}{29.48}&\multicolumn{2}{|c|}{22.53}&26.55 \\
		\hline
		ConvexOpt\cite{32}  &10.08&	11.63&	11.42&	14.58	&7.22&	6.17&10.28				 \\
		\hline
		Siam-Net\cite{2}&	13.21&	14.89&	8.77&	13.48&	8.38&	6.01&	10.07	\\
		\hline
		pseudo-siam\cite{2} &	12.64&	12.5&	12.87&	10.35&	5.44&	3.93&	9.62\\
		\hline
		2-ch Net\cite{2}&	6.04&	7&	6.05&	8.59&	3.05&	3.03&	5.63\\
		\hline
		Match-Net\cite{18}&11&13.58&8.84&13.02&7.7&4.75&	9.82\\
		\hline
		Match-Net\cite{18} &	\multirow{2}[2]{*}{8.39}&	\multirow{2}[2]{*}{10.88}&	\multirow{2}[2]{*}{6.90}&\multirow{2}[2]{*}{10.77}&	\multirow{2}[2]{*}{5.67}&\multirow{2}[2]{*}{3.87}&\multirow{2}[2]{*}{7.75}\\
		(no bottleneck)& & & & & & \\
		\hline
		PN-Net\cite{19}&7.21&	8.99&	8.13&	9.65&	4.23	&3.71&	6.98\\
		\hline
		CC-Net&	7.2&	9.2&	7.48&	8.4&	3.6&3.9&	6.63\\
		\hline
		\textbf{CT-Net}&	\textbf{5.5}&\textbf{6.57}	&	\textbf{5.47}&\textbf{5.25}	&\textbf{2.35}&\textbf{2.1}&\textbf{4.54}\\
		\hline
	\end{tabular}
	
	\caption{Test results on the Photo-Tour data set. CC-Net and CT-Net represent complex channel net and complex triple net respectively.  We reported the optimal results using bold Numbers.By comparison, Complex Triple Net gets state-of-the art results.}
	
\end{table*}

\subsection{Complex Triple Net}

The complex triple net consists of complex feature module and metric module. The main function of it is to learn the descriptor of the image. As shown in Figure 5(b) ,in the training step, the inputs of complex triple net are a triple of images. They input from different branches. The weights of the first convolutional layer and complex feature module of the three branches are shared with each other.

In this model, the descriptor can be obtained by sending the output of complex feature module to the metric module to reduce the dimension. We add the \( CL2\_norm\)(equation 10) at the end. According to the distance between the complex-valued descriptors obtained by using equation (7), we can increase the negative distance and reduce the positive distance by minimizing the loss function which is described equation (8).

In the test step, each image is directly converted to a descriptor. We can analyze the similarity of images by comparing the distances between descriptors, as shown in Figure 5(c)

The structure of convolution layer and complex feature module of two networks are the same.  During the training step, complex channel net has a faster training speed than complex triple net. However, it cannot give a representation of the image in the embedded space. Complex triple net can embed images into complex spaces. When performing image classification, we can quickly get results through a simple fully connected network, KNN, SVM and other classification algorithms.

\begin{table}
	\caption{This table shows the structure of CCN and CTN. In the conv/complex conv layer, the parameters given represent convolution kernel size, channel number and activation function.The parameters given in FC/Complex FC layer are the  number of output units and activation function}
	\resizebox{85mm}{45mm}{
		\centering
		\begin{tabular}{|c|c|c|}
			
			\hline
			
			&	CCN&	CTN\\
			\hline conv&	3*3,32,relu	&3*3,32,relu\\
			\hline Complex-BN&	Momentum=0.9&	Momentum=0.9\\
			\hline Complex-conv&	3*3,32,complex-relu&	3*3,32,complex-relu\\
			\hline Complex-BN&	Momentum=0.9&	Momentum=0.9\\
			\hline Complex-conv&	3*3,32,complex-relu&	3*3,32,complex-relu\\
			\hline pool&	complex\_pool&	complex\_pool\\
			\hline Complex-BN&	Momentum=0.9&	Momentum=0.9\\
			\hline Complex-conv&	3*3,64,complex-relu&	3*3,64,complex-relu\\
			\hline Complex-BN&	Momentum=0.9&	Momentum=0.9\\
			\hline Complex-conv&	3*3,64,complex-relu&	3*3,64,complex-relu\\
			\hline pool&	complex\_pool&	complex\_pool\\
			\hline Complex-BN&	Momentum=0.9&	Momentum=0.9\\
			\hline Complex-conv&	3*3,128,complex-relu&	3*3,128,complex-relu\\
			\hline Complex-BN&	Momentum=0.9&	Momentum=0.9\\
			\hline Complex-conv&	3*3,128&	3*3,128\\
			\hline pool&	complex\_pool&	complex\_pool\\
			\hline FC/Complex FC&	4096, relu	&128,Cl2-norm\\
			\hline FC&	4096, relu	&/\\
			\hline FC&	4096, relu	&/\\
			\hline FC&	1,sigmoid	&/\\
			\hline 
		\end{tabular}
	}

\end{table}

\begin{figure*}
	\begin{center}
		\includegraphics[width=0.8\linewidth]{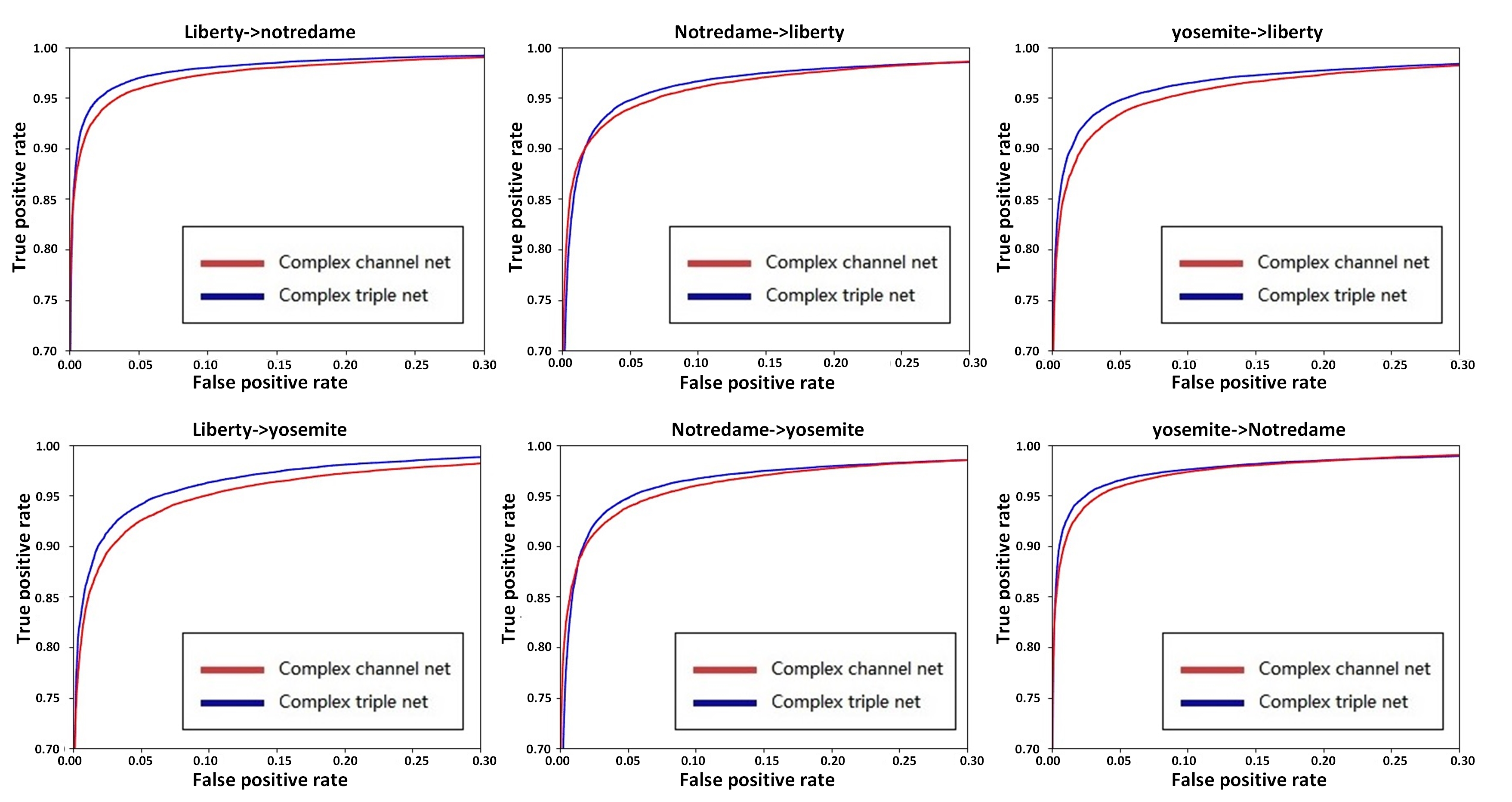}
	\end{center}
	\caption{ROC curves for our models on the local image patches benchmark.The red line represents complex channel net and the blue line represents complex triple net.
	}
	\label{fig:short}
\end{figure*}


\section{Experiment}

Photo-Tour dataset is the most commonly used dataset for patch verification task, our experiments are also based on it. In order to further illustrate the generalization ability of the model, we present the effect evaluation of proposed models on the HPatches dataset.

In this section, we provide the experimental details and effect evaluation of image matching. In section 4.1, we introduce the settings of parameter. Photo-Tour and HPatches dataset are introduced respectively in sections 4.2 and 4.3. The details of ablation study can be found in section 4.4. And in Section 4.5, we give the discussion and analysis the results of experiment.

\subsection{Experimental setup}

Complex channel Net, as described in section 3.4, uses an end-to-end training method to directly learn a similarity function with input as two pictures and output as a probability value. Complex channel net is trained with Adam optimization function. We set the learning rate to 1e-3, and each batch has a size of 128. Mean square error(MSE) as a loss function of the network.

Complex metric net, as described in section 3.5, determines the similarity of the image based on the descriptor distance. Although it is slower to train than complex channel net, it is able to give descriptors of different images and measure distances of different images.

In Complex metric net, we use the loss function proposed in this paper. The image descriptors are learned in the complex space, by optimizing the distance among the triples descriptors. Similar to the complex channel net, we set the learning rate to 1e-3, and each batch has a size of 128, using Adam optimization function.

\subsection{Photo-Tour dataset}
Photo-Tour dataset was referenced in the paper \cite{PTdataset},  which has been widely used in image matching and descriptor learning. The dataset consists of three subsets totaling more than 1.5 million patches. Each subset contains multiple sets of similar images. It is used by many researchers as a standard benchmark dataset to test the effects of metric learning \cite{2, 18, 19}.

The three subsets of it, names Notredame, Liberty and Yosemite. For each subset, the author gives 100k, 200k and 500k matching pairs, and the labels of them.  We used 500K pairs given by the author of the dataset as the training set, and the 100K patch pairs of the other subsets as the test to verify the effect of our model.

The test results of our proposed architectures on the Photo Tour are shown in table 1. The structure of Complex channel net (CCN) and Complex triple net (CTN) is described in detail in table 2. We report the false positive rate at 95\% recall (FPR95) on each of the six combinations of training and test sets, as well as the mean across all combinations. Complex triple net gets the competitive results. The ROC curve of our test is shown in Figure 6.

\subsection{HPatches dataset}

For further verifying the generalization ability of the experiment, we use the most current dataset named HPatches\cite{HPatches}. It’s a novel benchmark for evaluating local image descriptors. The dataset consists of 116 sequences of 6 images. We split the dataset into 5 parts: a, b, c, view and illum, according to the \emph{splits.json}, which is provided by the authors of HPatches.

In this paper, a, b and c are taken as the training data, and view and illum as the test data. We randomly sample data in a, b and c to construct the triple patch pairs in the training step. As with the Photo Tour dataset, we build 500K pairs as the training set and 100K pairs as the test set. We also trained CCN and CTN on the HPatches dataset and tested the effect on the Photo-Tour dataset. The test results are listed in table 3.

\begin{table}
	\setlength{\abovecaptionskip}{0pt}
	\setlength{\belowcaptionskip}{10pt}
	\caption{We train our model on the HPatches dataset and test it on the Photo-Tour dataset. The values of FPR95 for the model are recorded in the table.We introduce the results of L2-net, and by comparison, our performance gets the competitive results.}\centering
	
	\resizebox{85mm}{15mm}{
		\centering
		
		\begin{tabular}{c|c|c|c}
			
			\hline
			Train & \multicolumn{3}{c}{HPatches} \\
			\hline
			Test & Liberary&Notredame&Yosemite\\
			\hline
			L2-Net\cite{L2}&4.16&1.54&\textbf{4.41}\\
			\hline
			CCN&7.65&3.68&6.8\\
			\hline
			CTN&\textbf{4.10}&\textbf{1.32}&4.68\\
			\hline
			
		\end{tabular}
	}

\end{table}

\subsection{Ablation study}
In order to compare and verify the proposed method, we use simplified model to evaluate the effect. By comparing the parameters before and after the model simplification, the importance of the modified parameters can be evaluated.

We keep the original architecture of CCN and CTN, for verifying whether the complex-valued weights play a key role in the model. The difference is that the new models consist entirely of real numbers, denoted as Real channel network(RCN) and Real triple network(RTN) respectively. Since the loss function of complex-valued network is improved through PNSoft Loss, we use PNSoft as the loss of RCN and RTN in the training step.

We keep the original architecture of CCN and CTN, for verifying whether the complex-valued weights play a key role in the model. The difference is that the new models consist entirely of real numbers, denoted as Real channel network(RCN) and Real triple network(RTN) respectively. Since the loss function is improved through PNSoft Loss, we use it as the loss of RCN and RTN in the training step. The test set is the same as in section 4.3. The experimental results are recorded in figure 7.

\subsection{Discussion and analysis}

We propose two models names CCN and CTN, and verify the competitive performance of the two models in patch verification task through photo-tour dataset and HPatches dataset. In addition, through the ablation study, we analyze that the complex number certainly plays an important role in patch verification task.

Through the above experimental results, it can be found that although they are all the complex-valued network, the performance of CTN is significantly better than that of CCN. When building the network, we intentionally make the structure of the two models as similar as possible. But there is still a certain gap in their performance. We analysis that the main reason is that the essence of CTN is the learning of descriptors, while the essence of CCN is the learning of image similarity.

In the training step of CTN, images are converted into complex-valued descriptors, and the descriptors of different images are separated from each other as far as possible in the mapping space through the given loss function. By contrast, the similar images are close to each other as far as possible. In our view, Complex-valued vector can depict the detail information of the image deeply. Furthermore, the proposed measurement method of complex-valued vector can express the similarities and differences of images more clearly.

In the training step of CCN, it is possible that we fail to find a proper function to map the real image to the complex image. Simply setting the imaginary part to be initialized from zero may lose a lot of important information in the complex field. Therefore, the complex-valued network can not completely depict the image details when learning the image similarity directly. Although CTN also initializes the imaginary part from 0, it may be possible to mitigate this by optimizing the measure distance in the complex domain.

\begin{figure}[t]
	\begin{center}
		\includegraphics[width=1\linewidth]{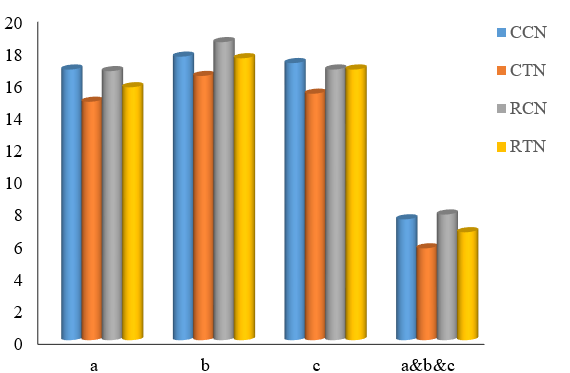}
	\end{center}
	\caption{FPR95 histogram of four models. The horizontal axis represents the training model on different data sets, and the vertical axis represents the FPR95 value of the model output in the test set.}
	\label{fig:long}
	\label{fig:onecol}
\end{figure}

\section{Conclusion}
We propose a new approach to learn descriptor in complex domain. Compared to the image descriptors in the real domain, we believe that the complex domain can describe more detailed information. In this paper, we’ve developed a way to measure the distance of the complex-valued vector, and we’ve improved the loss function of the PN-Net in complex field. We extend the work in \cite{10}, using complex-block to build the complex-valued residual network. In the end, our model achieved the best result so far. 

After extracting features through complex convolutional
neural network, we also tried to use siamese networks to
process the feature values of the real part and the imaginary part respectively and combine them in a fully connected layer, which is consistent with the idea of 2-ch net \cite{2}.

These work show that, complex-valued networks is competitive in descriptor learning. Complex-valued descriptors are more suitable for metric learning. However, it’s important to point out, we still lack of the methods to accelerate complex-valued network training, relative to the real-valued neural networks, complex neural network training speed remains to be improved.

{\small
	\bibliographystyle{ieee}
	\bibliography{egbib}
}
\end{document}